\documentclass[pdflatex,sn-mathphys-num]{sn-jnl}% Math and Physical Sciences Numbered Reference Style
\usepackage{graphicx}%
\usepackage{multirow}%
\usepackage{amsmath,amssymb,amsfonts}%
\usepackage{amsthm}%
\usepackage{mathrsfs}%
\usepackage[title]{appendix}%
\usepackage{xcolor}%
\usepackage{textcomp}%
\usepackage{manyfoot}%
\usepackage{booktabs}%
\usepackage{algorithm}%
\usepackage{algorithmicx}%
\usepackage{algpseudocode}%
\usepackage{listings}%
\usepackage[shortcuts,acronym]{glossaries}%
\usepackage{ulem}
\usepackage{subcaption}
\DeclareGraphicsExtensions{.pdf, .png}                      %  will firstly be searched and used (Prerequisites -> no 
\makeglossaries
\glsdisablehyper
\newcommand{\acmath}[1]{\text{\glsentryshort{#1}}}%

\newacronym{ai}{AI}{Artificial Intelligence}
\newacronym{hlas}{HLAS}{High-Level Autonomy System}
\newacronym{aeb}{AEB}{Autonomous Emergency Braking}
\newacronym{aebs}{AEBS}{Autonomous Emergency Braking System}
\newacronym{fcw}{FCW}{Forward Collision Warning}
\newacronym{lidar}{LiDAR}{Light Detection and Ranging}
\newacronym{fov}{FOV}{Field of View}
\newacronym{nefimmore}{NeFiMMORE}{Near-Field Monitoring System MORE}
\newacronym{more}{MORE}{Munich Mobility Research Campus}
\newacronym{pmcm}{PMCM}{Predictive Motion Corridor Monitoring}
\newacronym{gdeta}{GDETA}{Gated Dual-Evidence Threat Assessment}
\newacronym{ttc}{TTC}{Time-to-Collision}
\newacronym{ttb}{TTB}{Time-to-Brake}
\newacronym{tts}{TTS}{Time-to-Steer}
\newacronym{th}{TH}{Time Headway}
\newacronym{ttd}{TTD}{Time-to-Disappear}
\newacronym{btn}{BTN}{Brake Threat Number}
\newacronym{drac}{DRAC}{Deceleration Rate to Avoid a Collision}
\newacronym{thw}{THW}{Time Headway}
\newacronym{lfb}{LFB}{Last-Feasible-Brake}
\newacronym{ri}{RI}{Risk Index}
\newacronym{aabb}{AABB}{Axis-Aligned Bounding Box}
\newacronym{fnr}{FNR}{False Negative Rate}
\newacronym{fpr}{FPR}{False Positive Rate}
\newacronym{tpr}{TPR}{True Positive Rate}
\newacronym{tp}{TP}{True Positive}
\newacronym{fn}{FN}{False Negative}
\newacronym{fp}{FP}{False Positive}
\newacronym{tn}{TN}{True Negative}
\newacronym{cr}{CR}{Collision Rate}
\newacronym{rss}{RSS}{Responsibility-Sensitive Safety}
\newacronym{nbr}{NBR}{Nuisance Braking Rate}
\newacronym{adas}{ADAS}{Advanced Driver Assistance System}
\newacronym{sotif}{SOTIF}{Safety of the Intended Functionality}
\newacronym{nlp}{NLP}{Nonlinear Programming}
\newacronym{ttr}{TTR}{Time-to-React}
\newacronym{ttf}{TTF}{Time-to-Front}
\newacronym{imm}{IMM}{Interacting Multiple Model}
\newacronym{mdp}{MDP}{Markov Decision Process}
\theoremstyle{thmstyleone}%
\theoremstyle{thmstyletwo}%

\theoremstyle{thmstylethree}%

\begin{document}

\title[Article Title]{Corridor-Based Monitoring and Emergency Braking Decision for Automated Vehicle Safety}


\author[1]{\fnm{Junnan} \sur{Pan}}\email{junnan.pan@unibw.de}
\author*[1]{\fnm{Vladislav} \sur{Nenchev}}\email{vladislav.nenchev@unibw.de}
\author[1]{\fnm{Mohak} \sur{Mansharamani}}\email{mohak.mansharamani@unibw.de}
\author[1]{\fnm{Prodromos} \sur{Sotiriadis}}\email{prodromos.sotiriadis@unibw.de}
\author[1]{\fnm{Ferdinand} \sur{Englberger}}\email{ferdinand.englberger@unibw.de}

\affil[1]{\orgdiv{Department of Electrical and Computer Engineering}, \orgname{University of the Bundeswehr Munich}, \orgaddress{\street{Werner-Heisenberg-Weg}, \city{Neubiberg}, \postcode{85579}, \country{Germany}}}

\abstract{This paper addresses emergency-braking decisions for a redundant safety channel in automated vehicles. The channel is responsible for identifiying hazards from sensor data and determining whether braking is necessary, given the constrained computational resources of embedded hardware. We present an emergency-braking approach that uses a predicted motion corridor to filter \ac{lidar} measurements and then decide whether the ego vehicle must brake. \ac{pmcm} constructs a curvature-aware boundary from the planned path and vehicle geometry so that the monitored region contains the full swept body on both straight and curved paths. \ac{gdeta} then decides whether the ego vehicle must command an emergency brake for objects inside this boundary, scoring geometric occupancy and temporal urgency independently. Hard gates permit braking when either source is conclusive; soft-OR arbitration combines partial evidence when neither gate is active. Consecutive-scan confirmation reduces sensitivity to transient measurement noise. Corridor coverage is evaluated in controlled collision simulations and recorded campus driving. Braking decisions are evaluated in Monte Carlo simulations of lead-vehicle braking and transverse-crossing encounters.
}




\maketitle

\section{Introduction}\label{sec:intro}
Modern \ac{adas} stacks increasingly rely on \ac{ai}, especially deep neural networks, to raise nominal perception performance \cite{Shivanna2024Object}. These models are not easy to inspect and may produce detections that are hard to explain \cite{zablocki2022explainability}, or they may degrade on rare, long-tail traffic situations that are poorly represented in the training data \cite{Abrecht2024Deep, Peng2024Curse}. A second channel, with its own sensor, decision rules, and authority to stop the vehicle, is an established redundancy when the main autonomy stack misses a hazard or leaves too little time to avoid a collision~\cite{hanselaar2024safety,pan2024independent}.

This paper studies such a second channel that runs on low-cost hardware, such as microcontrollers, and that can support a safety argument. The monitor takes a trajectory planned by the \ac{hlas} and raw scans from a single \ac{lidar}, and at each scan decides whether an emergency stop is required within a bounded monitoring region.

For such resource-limited monitoring systems, two questions follow. First, how can the monitoring area be restricted to the traffic participants that may collide with the ego vehicle as it follows the planned path? Second, once those participants occupy that area, how can the available object measurements and the known ego motion be used to decide an emergency brake command?

Last-feasible-brake rules determine when to stop based on objects that have already been tracked~\cite{kaempchen2009situation,brannstrom2010model}, but they generally assume that geometric relevance has already been established. Where an explicit monitoring region is used, it is often a coarse keep-out mask~\cite{pan2024independent,gutjahr2014automatic}. On a tight turn, such a mask may be too narrow along the outer edge, so an obstacle enters the monitored region only after the available braking distance has been consumed. The same mask may be too wide beside the path, so curbs or parked vehicles appear relevant.

We use the same predicted motion corridor to retain \ac{lidar} returns within the swept body and to decide whether the ego vehicle must command an emergency brake for the resulting objects. The geometric basis is \ac{pmcm}, a curvature-aware monitoring boundary. The planned rear-axle path is expressed in a curvilinear frame, the four body corners are propagated along it, and sliding extrema of their tracks form an envelope that contains the swept body at any curvature. \ac{gdeta} then scores geometric occupancy and temporal urgency independently on this envelope, so either source can command a stop when its evidence is conclusive.

Keeping the two evidence sources separate lets \ac{gdeta} use the information the tracker already has. Object position is available before velocity is reliable, so the geometric evidence can command braking once the object occupies the emergency envelope for the required number of scans, or immediately if the object is already close and blocks enough of the corridor. After a valid motion estimate exists, the temporal evidence checks whether the object will still overlap the ego width when the ego reaches that longitudinal position, and whether braking has become urgent. The temporal condition must persist over consecutive scans before a command is issued, so a brief predicted overlap or one noisy measurement does not stop the vehicle.

Evaluation starts with corridor coverage and then isolates the brake rule. A controlled collision series places a static obstacle at the outer corridor edge across a range of speeds and turning radii. Among the three masks considered, the \ac{pmcm} envelope is the only one that both detects the obstacle in every trial and prevents collisions in all tested cases after braking. The parallelogram mask detects the obstacle too late, while the two-point strip often fails to detect it. The \ac{gdeta} threat assessment decision rule is then evaluated independently in two Monte Carlo campaigns. An established lead-vehicle case tests temporal assessment with the geometric gate suppressed by a confirmed track. A transverse-crossing case tests geometric intervention during the first scans after a sudden corridor entry, before a reliable motion estimate and tracker lock are available.

This article extends our previous conference publication~\cite{pan2026enhancing}, which introduced the curvature-aware \ac{pmcm} method and evaluated it in a case study. The present article reformulates \ac{pmcm} and extends it with:
\begin{enumerate}
    \item \ac{gdeta}, a deterministic threat-assessment rule that scores geometric occupancy and temporal urgency independently, admits a hard gate on either source, and uses soft-OR arbitration when no gate applies.
    \item An evaluation that separates corridor coverage from braking decisions and examines \ac{gdeta} in established lead-vehicle braking and transverse-crossing encounters.
\end{enumerate}

\section{Related Work}\label{sec2}
This section reviews emergency braking and independent safety channels, corridor and envelope monitoring, and threat-assessment metrics for an embedded brake rule.
\subsection{Emergency Braking and Independent Safety Channels}
\ac{aeb} systems estimate collision risk from nearby objects and apply a brake command when necessary~\cite{yang2022systematic}. Production algorithms are mainly longitudinal and work on an object list: last-feasible braking when a collision is unavoidable given vehicle dimensions and orientation~\cite{kaempchen2009situation}, model-based remaining-maneuver sets that also include steering and acceleration~\cite{brannstrom2010model}, and later timing rules that delay full brake for comfort~\cite{han2024adaptive}. They are built as a primary \ac{adas} function. The input is a tracked object list; the planned automated trajectory is not part of the monitor.

Emergency intervention also appears as an independent safety channel beside a higher-level stack. Hanselaar~et al.~\cite{hanselaar2024safety} propose a multi-channel Safety Shell that arbitrates among redundant functions under a \ac{sotif} argument. Pek~et al.~\cite{pek2020using,pek2021fail} execute an intended trajectory only while a collision-free fail-safe trajectory remains available under set-based occupancies of other road users, using both steering and braking. A \ac{lidar}-based \ac{aeb} system for a heavy vehicle uses sparse flash returns~\cite{kumar2024autonomous}. It connects the sensor to the brake command and selects relevant targets, still as a primary \ac{aeb} function.

Althoff and Dolan verify candidate vehicle motion online against reachable occupancies~\cite{althoff2014online}. Gruber and Althoff return conservative verification results under varying computational load with an anytime refinement scheme~\cite{gruber2018anytime}. Magdici and Althoff keep a verified fail-safe motion behind the intended trajectory~\cite{magdici2016failsafe}, and Sch\"urmann~et al.\ formally check whether planned motions remain dynamically drivable~\cite{schuermann2017drivability}. vom Dorff~et al.\ separate a safety observer from the nominal planner and buffer short trajectories that end in a safe stop~\cite{vomdorff2020failsafe}. These methods check planned motion at runtime and keep a safe stop available.

\subsection{Corridor and Envelope Monitoring}
Geometric monitoring regions can serve as the collision-check region for braking. Althoff~et al.~\cite{Althoff2009Model-Based} predict the stochastic road occupancy of surrounding vehicles and integrate crash probability along a given ego trajectory, so the monitored region is a time-varying occupancy density. Lee and Kum~\cite{Lee2019Collision} and Shen~et al.~\cite{Shen2020Collision-Free} construct a predictive occupancy map from the relative position, velocity and acceleration of nearby traffic, score a discrete set of acceleration candidates on that map, and trigger collision mitigation from the same occupancy.

On curved roads, a fixed or straight-ahead region assigns the wrong objects to the path. Lee~et al.~\cite{Lee2019Study} convert host and target states into curvilinear, road-aligned coordinates before computing the relative distance and time-to-collision that trigger \ac{aeb}. Han~et al.~\cite{Han2025Adaptive} reshape the region of interest online from steering angle and longitudinal velocity, and adapt its range under a worst-case road geometry without an explicit lane or curvature estimate. That adaptation reduces false detections of roadside objects such as guardrails and raises detection of genuine in-path threats. Collision checking can also use a free-space contour: Gao~et al.~\cite{Gao2022Deformable} represent instantaneous free space as a Doppler-deformable polygon and check collisions against that polygon instead of a dense occupancy grid.

\subsection{Threat Assessment}
Dahl~et al.~\cite{dahl2019collision} and Westhofen~et al.~\cite{westhofen2023criticality} group collision metrics into single-behavior scalars (\ac{ttc}, \ac{th}, \ac{ttb}, \ac{btn}, \ac{ttr}), optimization and reachability, probabilistic frameworks, and learning-based scores. The monitor uses the braking-only scalars \ac{ttc}, \ac{ttb} and \ac{btn}, which an embedded rule can evaluate deterministically.

Jansson and Gustafsson~\cite{jansson2008framework} give a Bayesian collision-avoidance decision from required acceleration, the quantity later called \ac{btn}. Br\"annstr\"om~et al.~\cite{brannstrom2010model} compute an analytical remaining set of steer, brake and accelerate maneuvers for a rectangular host and a polygonal object, and brake when that set is empty. Kaempchen~et al.~\cite{kaempchen2009situation} trigger when a collision is unavoidable over all physically possible trajectories. Gelso and Sj\"oberg~\cite{gelso2017consistent} use \ac{ttb} and \ac{btn} together, with road geometry to predict lead-vehicle deceleration, to suppress false alarms at junctions on euroFOT data.

Reachable sets and constraint satisfaction bound last-chance timing more tightly than a scalar \ac{ttb}. Falcone~et al.~\cite{falcone2011predictive} treat threat as constraint satisfaction and receding-horizon safe sets for road departure. S\"ontges, Koschi and Althoff~\cite{sontges2018worst} over-approximate \ac{ttr} with reachable sets; an empty set means no evasive maneuver remains. S\'anchez Morales~et al.~\cite{sanchez2019parallel} evaluate millions of ego/object pose hypotheses on a Drive PX~2 in about $21\,\mathrm{ms}$.

Xu, Zhao and Wang~\cite{xu2020integrated} synthesize \ac{ttc}, time-headway and a \acl{ttf} term after \acl{imm} prediction, then search an \acl{mdp} for a trajectory; the fused score is a planner cost. \ac{rss}~\cite{shalevshwartz2017formal} specifies formal longitudinal and lateral duty-of-care distances. The geometric emergency envelope here is the planned corridor limited to a kinematic delay-plus-braking depth.

Hillenbrand~et al.\ organize collision mitigation as multilevel situation assessment and decision-making, and make the intervention trade-offs explicit~\cite{hillenbrand2006multilevel}. Eidehall and Petersson estimate collision threat by Monte Carlo sampling of uncertain road scenes~\cite{eidehall2008statistical}. Schreier, Willert and Adamy couple maneuver recognition, trajectory prediction and criticality assessment in arbitrary road layouts~\cite{schreier2016integrated}. Kim and Kum combine lane-conditioned motion probabilities with trajectory-pair \ac{ttc} into a collision-risk map over candidate ego paths~\cite{kim2018lane}. These methods use prediction distributions, maneuver hypotheses or candidate trajectories.

\section{System Overview}\label{sec3}
\begin{figure} [t]
    \centering
    \includegraphics[width=0.7\columnwidth]{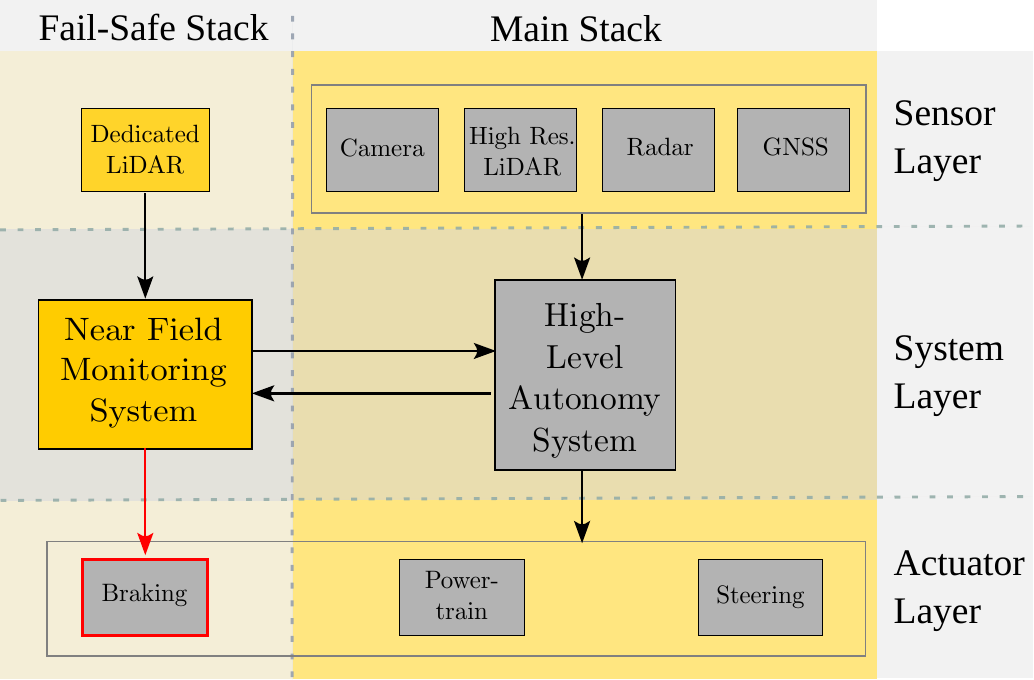}
    \caption{Automated vehicle technical stack. The fail-safe stack is \ac{nefimmore} with a dedicated \ac{lidar} and brake authority. The main stack is the \ac{hlas} with the full sensor suite, steering, and powertrain. The only runtime coupling is the planned trajectory toward the monitor and a warning plus brake request back to the vehicle \cite{pan2026enhancing}.}
    \label{fig:sys:architecture}
\end{figure}
\ac{nefimmore} is the dedicated near-field monitoring system of the \ac{more} experimental vehicle. It uses its own 3D \ac{lidar} and an embedded host. The hardware platform architecture is given in~\cite{pan2023collision,pan2024independent}.
\subsection{System Description}\label{subsec3.1}

\begin{figure} [t]
    \centering
    \includegraphics[width=0.85\columnwidth]{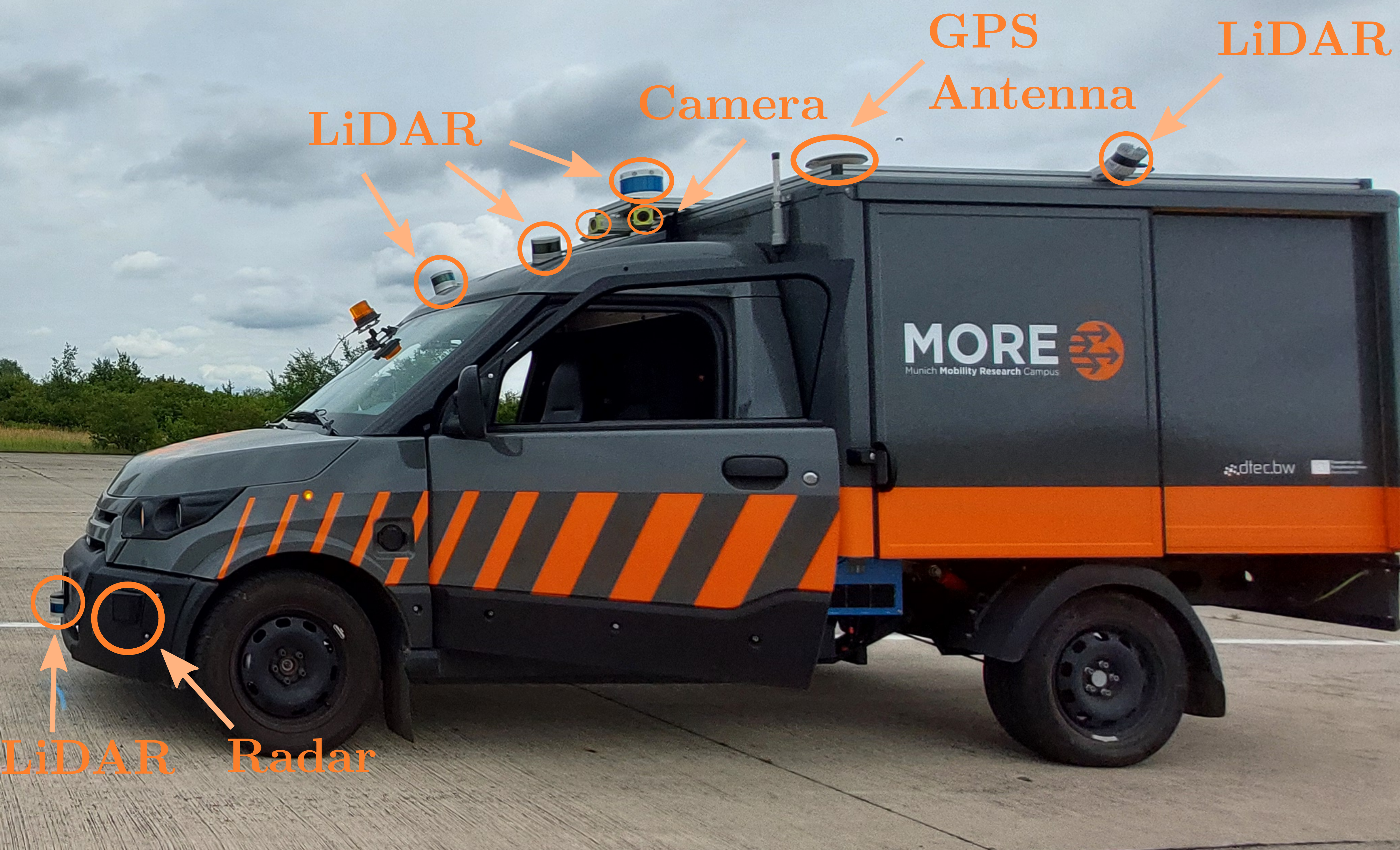}
    \caption{Experimental vehicle sensor setup. Our \ac{nefimmore} system uses a low-mounted front 3D \ac{lidar} to perceive the near-field environment.}
    \label{fig:sys:sensor_setup}
\end{figure}

Figure~\ref{fig:sys:architecture} shows the autonomous vehicle technical stack. The platform is split into a main stack and a fail-safe stack, each with a system layer, a sensor layer, and an actuator layer. The main stack is the \ac{hlas}: a high-performance computer with roof-mounted high-resolution \ac{lidar}, camera, radar, and GNSS (Figure~\ref{fig:sys:sensor_setup}). It plans a collision-free path and commands steering and the powertrain. The fail-safe stack is \ac{nefimmore}: an embedded host and a low-mounted dedicated \ac{lidar}, with authority to command the service brake. The stacks share two messages. The \ac{hlas} sends the planned trajectory. The monitor returns a warning when a relevant object occupies the predicted motion corridor, and a brake command when Sect.~\ref{subsec4.2} requires a stop. \ac{nefimmore} is the redundant last-resort channel: the \ac{hlas} remains the first actor on a conflict it can observe in time, and the monitor commands a stop for a sudden corridor hazard that the roof suite cannot process in time.

The roof \ac{lidar} can detect far traffic, but it may still miss a near-field object in the dead corner of the roof \ac{lidar}. From the planned path and the current speed, the monitor builds a predicted motion corridor whose length covers reaction plus braking and whose width follows the body sweep~\cite{pan2025reliability,pan2026enhancing}. A low-mounted front \ac{lidar} watches the volume near the ground inside that corridor.

  \begin{figure}[t!]
    \centering
    \includegraphics[width=\columnwidth]{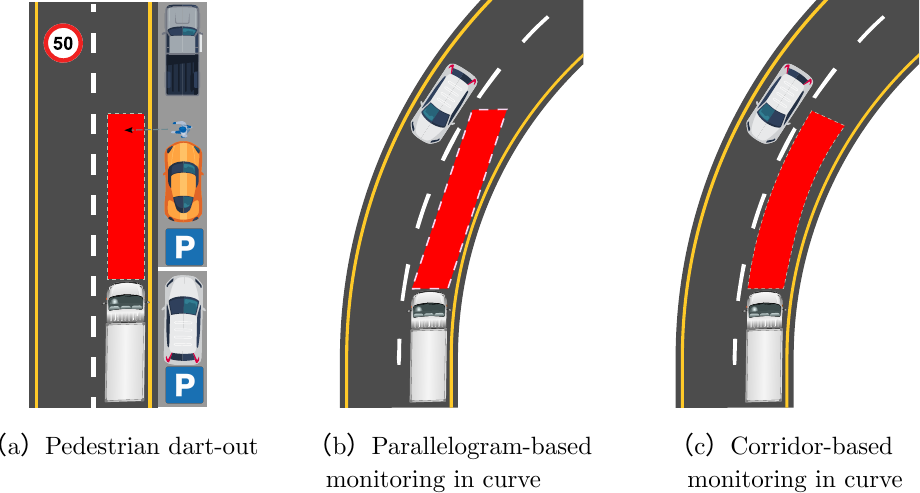}
    \caption{Predicted motion corridor (marked in red) in representative near-field scenes. Longitudinal depth follows the current speed; lateral bounds follow the planned body sweep \cite{pan2026enhancing}.}
    \label{fig:sys:monitor_scenes}
  \end{figure}

Figure~\ref{fig:sys:monitor_scenes} illustrates the monitoring corridor in representative scenes. The red region marks the keep-out area. A speed-dependent rectangle is sufficient on a straight road~\cite{pan2025reliability}; on a tight turn, however, the corridor must follow the geometric boundary of the ego vehicle. Section~\ref{subsec4.1} describes this construction.

\subsection{System Pipeline}\label{subsec3.2}

\begin{figure}[ht]
    \centering
    \includegraphics[width=1.0\textwidth]{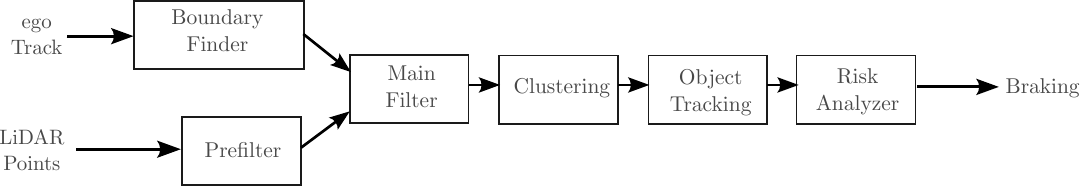}
    \caption{Overview of the runtime processing pipeline implemented in \ac{nefimmore}.}
    \label{fig:pipeline_structure}
\end{figure}
Figure~\ref{fig:pipeline_structure} shows the runtime pipeline from corridor monitoring to the brake decision. Two inputs enter in parallel: the planned ego track from the \ac{hlas} and the raw sensor input from the dedicated \ac{lidar}. A prefilter and a main filter discard \ac{lidar} samples outside the region of interest, which reduces the memory and compute load of processing the raw point cloud. The retained points are grouped into objects, associated across scans, and passed to the risk analyzer. The boundary finder and the main filter implement \ac{pmcm} (Sect.~\ref{subsec4.1}). The risk analyzer implements \ac{gdeta} (Sect.~\ref{subsec4.2}).

The boundary finder builds the monitoring corridor whose length follows the current speed and whose width follows the predicted body sweep. The prefilter crops the \ac{lidar} sweep to a near-field volume in the sensor frame, so returns outside that volume never reach the corridor test. The main filter then maps the corridor into the \ac{lidar} frame and retains the \ac{lidar} returns that lie inside it.

Clustering groups those points by Euclidean region growing~\cite{rusu20113d}. Object tracking associates clusters across scans by Hungarian assignment of centroids~\cite{kuhn1955hungarian}, keeps a stable identifier, and estimates motion. The footprint and the motion are then projected onto the planned path, so each track is expressed in the corridor coordinates that \ac{gdeta} uses. The risk analyzer receives this object list together with the ego motion. Geometric occupancy of the emergency envelope and temporal braking urgency are scored independently, and either source may command a stop. The pipeline output is a binary hard-brake request.

\section{Proposed Methodology}\label{sec4}
\subsection{Curve-aware Monitoring using \texorpdfstring{\ac{pmcm}}{PMCM}}\label{subsec4.1}

\ac{pmcm} maps the planned driving path of the \ac{hlas} and the ego vehicle geometry onto a curvature-aware region of interest, then retains only those \ac{lidar} returns that fall inside this region. The region follows the predicted body sweep, so it remains valid when path curvature changes. On a tight urban turn, a fixed-width or parallelogram mask is either too small, so it misses the outer part of the swept path, or too large, so it treats sidewalk, curb, or parked objects beside the turn as if the ego vehicle would drive through them. Clustering and tracking operate on corridor-resident points, and \ac{gdeta} in Sect.~\ref{subsec4.2} assesses only objects that already lie inside this monitored set.

The procedure has three stages. First, a motion corridor is obtained from sliding extrema of the vehicle-corner tracks. Second, that corridor is reduced to a braking-horizon, sensor-visible monitoring boundary. Third, each return is tested against azimuth-local triangles of this boundary.

\subsubsection{Motion-corridor Envelope}
\begin{figure}[th]
    \centering
    \includegraphics[width=0.85\textwidth]{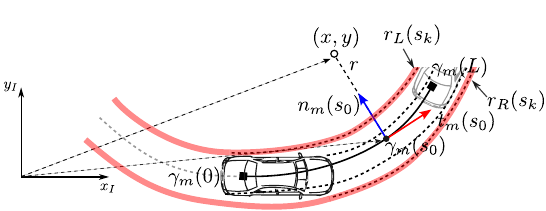}
    \caption{Predicted reference path $\gamma_m(s)$ in a curvilinear frame, together with the motion-corridor envelope obtained from the vehicle-corner tracks~\cite{pan2026enhancing}.}\label{fig:method:frenet_coordinate}
\end{figure}

A curvilinear frame is attached to the predicted reference track of the rear-axle midpoint, shown in Figure~\ref{fig:method:frenet_coordinate}. Let $\gamma_m:[0,L]\to\mathbb{R}^2$ be a $C^2$ planar curve parameterized by arc length $s$,
\begin{equation}
\gamma_m(s)=\begin{pmatrix}x_m(s)\\ y_m(s)\end{pmatrix},\label{eq:pmcm_gamma}
\end{equation}
with unit tangent $t_m(s)=\gamma_m'(s)$ and leftward unit normal
\begin{equation}
n_m(s)=\begin{pmatrix}-y_m'(s)\\ x_m'(s)\end{pmatrix}.\label{eq:pmcm_normal}
\end{equation}
A ground-plane point then has coordinates $(s,r)$ through $\mathbf{p}(s,r)=\gamma_m(s)+n_m(s)\,r$. The same corridor frame is later used by \ac{gdeta}, where the lateral coordinate is written $d$.

The vehicle body $\mathcal{R}$ is modeled as a rectangle
\begin{equation}
\mathcal{R}=\bigl\{(a,b)\in\mathbb{R}^2:\; -l_r\le a\le l_f,\; |b|\le w/2\bigr\},\label{eq:pmcm_rect}
\end{equation}
where $l_r$ and $l_f$ are the rearward and forward overhangs from the vehicle baselink and $w$ is the outer width. At pose $s_0$, a body point $(a,b)\in\mathcal{R}$ maps to $\mathbf{p}_{\mathrm{car}}(s_0,a,b)=\gamma_m(s_0)+a\,t_m(s_0)+b\,n_m(s_0)$. The swept set $S=\bigcup_{s_0\in[0,L]}(\gamma_m(s_0)+\mathcal{R}_{s_0})$ is the region that must be monitored. Lateral bounds $r_L(s)$ and $r_R(s)$ are required to satisfy
\begin{equation}
S\subseteq\bigl\{(s,r):\; r_R(s)\le r\le r_L(s)\bigr\}.\label{eq:pmcm_contain}
\end{equation}

Rather than forming this Minkowski sweep explicitly, the method tracks the four corner trajectories
$\{(s_i^{\mathrm{lf}},r_i^{\mathrm{lf}})\}$,
$\{(s_i^{\mathrm{lr}},r_i^{\mathrm{lr}})\}$,
$\{(s_i^{\mathrm{rf}},r_i^{\mathrm{rf}})\}$,
$\{(s_i^{\mathrm{rr}},r_i^{\mathrm{rr}})\}$,
where the superscripts denote the left front, left rear, right front, and right rear corners. The tracks are sampled on a common grid $s_k=k\Delta s$, where $\Delta s$ is the sampling interval. At $s_k$ the body occupies the interval $[s_k-l_r,s_k+l_f]$. With forward window length $F=\lceil l_f/\Delta s\rceil$ and backward window length $R=\lceil l_r/\Delta s\rceil$, the left and right envelopes are the sliding extrema
\begin{align}
r_L(s_k)
&=
\max\Bigl\{
\max_{j\in[k,k+F]} r_{\mathrm{lf}}(s_j),\;
\max_{j\in[k-R,k]} r_{\mathrm{lr}}(s_j)
\Bigr\},
\nonumber\\
r_R(s_k)
&=
\min\Bigl\{
\min_{j\in[k,k+F]} r_{\mathrm{rf}}(s_j),\;
\min_{j\in[k-R,k]} r_{\mathrm{rr}}(s_j)
\Bigr\}.\label{eq:pmcm_extrema}
\end{align}
Each sliding extremum is linear in the number of samples. The resulting polylines, drawn in red in Figure~\ref{fig:method:frenet_coordinate}, are the motion corridor used by all subsequent stages.

\subsubsection{Visible Monitoring Boundary}
\begin{figure}[t]
    \centering
    \includegraphics[width=1.0\textwidth]{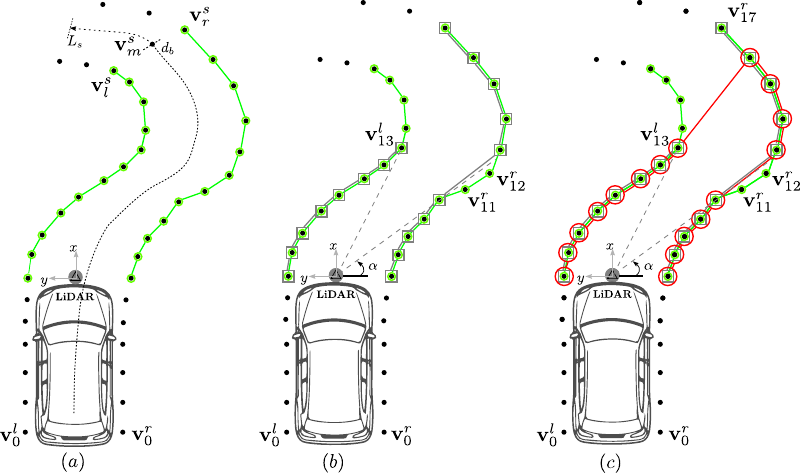}
    \caption{Construction of the visible monitoring boundary from a predicted motion corridor: (a)~boundary preconstruction; (b)~visible reduction; (c)~occlusion filtering~\cite{pan2026enhancing}.}\label{fig:method:boundary_finder_on_synthetic_corridor}
    \end{figure}

The full envelope is larger than the region that a front-center \ac{lidar} can observe before a stop. An operational monitoring boundary is therefore required to respect the braking distance, to contain only vertices that are visible from the sensor, and to use the available \ac{fov}. Computation is performed in the \ac{lidar} frame, whose $x$-axis points forward. The three reduction stages are shown in Figure~\ref{fig:method:boundary_finder_on_synthetic_corridor} and are described next.

\paragraph{Boundary preconstruction.}
Let $\mathcal{P}_{i}^{r}$ and $\mathcal{P}_{j}^{l}$ denote the discrete vertices along the right and left corridor boundaries from Eq.~\eqref{eq:pmcm_extrema}, now expressed in the \ac{lidar} frame. Only samples with $x\ge 0$ are kept; vertices behind the sensor are ignored. If a side has no discrete point at $x=0$, that intersection with the lateral axis is interpolated so the polygon is closed at the sensor. Let $\Delta_k=\|(x_{k+1},y_{k+1})-(x_k,y_k)\|_2$ be successive segment lengths along the predicted front-\ac{lidar} path and $L_s=\sum_{k=0}^{s-1}\Delta_k$ the cumulative length. The stopping index $s$ satisfies $L_s\le d_b<L_{s+1}$, where $d_b$ is the braking distance. The stopping pose $\mathbf{v}_{m}^{s}$ is interpolated between $\mathbf{v}_{s}$ and $\mathbf{v}_{s+1}$ so that the path length equals $d_b$, then projected onto both sides, giving $\mathbf{v}_{r}^{s}$ and $\mathbf{v}_{l}^{s}$. The preconstructed set
\begin{equation}
\mathcal{S}^{p}
=
\{\mathcal{P}_{i}^{r}:\; x\ge 0,\; i=0,\ldots,s-1\}
\cup
\{\mathbf{v}_{r}^{s}\}
\cup
\{\mathcal{P}_{j}^{l}:\; x\ge 0,\; j=0,\ldots,s-1\}
\cup
\{\mathbf{v}_{l}^{s}\}
\label{eq:pmcm_spre}
\end{equation}
is the primary forward, braking-limited corridor (green circles connected by line segments in the left panel of Figure~\ref{fig:method:boundary_finder_on_synthetic_corridor}).

\paragraph{Visible reduction.}
This step walks each ordered chain and retains only vertices that remain extremal in azimuth from the sensor origin $\mathbf{o}$, collecting $\widehat{\mathcal{S}}^{r}$ and $\widehat{\mathcal{S}}^{l}$. With $\alpha(\mathbf{v})=\operatorname{atan2}(y,x)+\pi/2$, the right accumulator stores the current maximum azimuth and the left accumulator the current minimum. Self-occluded samples do not update these extrema: in the center panel of Figure~\ref{fig:method:boundary_finder_on_synthetic_corridor}, $\mathbf{v}_{11}^{r}$, $\mathbf{v}_{12}^{r}$, and the vertices lying behind $\mathbf{v}_{13}^{l}$ are excluded.

\paragraph{Occlusion filtering.}
This step removes candidates that are hidden by the opposite corridor. In the right panel of Figure~\ref{fig:method:boundary_finder_on_synthetic_corridor}, $\mathbf{v}_{17}^{r}$ is occluded by $\mathbf{v}_{13}^{l}$ and is discarded. Let $\widehat{\mathcal{S}}_{\mathrm{trim}}^{r}$ and $\widehat{\mathcal{S}}_{\mathrm{trim}}^{l}$ be the vertices of $\widehat{\mathcal{S}}^{r}$ and $\widehat{\mathcal{S}}^{l}$ that remain after this opposite-corridor test. The surviving vertices of both sides are merged into a single sequence ordered by increasing azimuth,
\begin{equation}
(\mathbf{v}_{1},\ldots,\mathbf{v}_{M})
:=\operatorname{sort}_{\alpha}\bigl(\widehat{\mathcal{S}}_{\mathrm{trim}}^{r}\cup\widehat{\mathcal{S}}_{\mathrm{trim}}^{l}\bigr),\label{eq:pmcm_sort}
\end{equation}
which is the open polygonal chain used for point inclusion in the next step.

\begin{figure}[t]
    \centering
    \includegraphics[width=0.4\columnwidth]{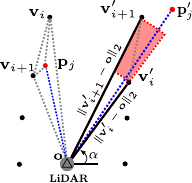}
    \caption{Local azimuth window of two successive visible vertices. Point $\mathbf{p}_j$ lies inside the filtering triangle; $\mathbf{p}_j'$ lies outside the admissible range~\cite{pan2026enhancing}.}\label{fig:method:window_points_filter}
    \end{figure}

\subsubsection{Azimuth-window Point Filter}
Consecutive vertices $\mathbf{v}_{i}\mathbf{v}_{i+1}$ of Eq.~\eqref{eq:pmcm_sort} form a local window. A 3D \ac{lidar} return is first projected onto the 2D plane. The resulting $\mathbf{p}_j$ is retained if its azimuth lies in $[\alpha(\mathbf{v}_i),\alpha(\mathbf{v}_{i+1}))$ and it belongs to the closed triangle $\overline{\triangle(\mathbf{o},\mathbf{v}_i,\mathbf{v}_{i+1})}$. Membership is decided by consistent orientation of the three edge cross products
\begin{align}
s_1&=(\mathbf{v}_{i+1}-\mathbf{o})\times(\mathbf{p}_j-\mathbf{o}),\nonumber\\
s_2&=(\mathbf{v}_{i+1}-\mathbf{v}_{i})\times(\mathbf{p}_j-\mathbf{v}_{i}),\nonumber\\
s_3&=(\mathbf{o}-\mathbf{v}_{i+1})\times(\mathbf{p}_j-\mathbf{v}_{i+1}).\label{eq:pmcm_cross}
\end{align}

The orientation test holds when $s_1,s_2,s_3$ are all non-negative or all non-positive:

\begin{equation}\label{eq:cond}
    (s_1 \ge 0 \wedge s_2 \ge 0 \wedge s_3 \ge 0)\ \ \vee\ \ (s_1 \le 0 \wedge s_2 \le 0 \wedge s_3 \le 0)
\end{equation}

Figure~\ref{fig:method:window_points_filter} shows a return $\mathbf{p}_j$ inside the window and a return $\mathbf{p}_j'$ that fails it. The orientation test is expensive if applied to every 3D return, so a candidate is rejected unless it lies no farther than the more distant of the two window vertices
\begin{equation}
\|\mathbf{p}_j-\mathbf{o}\|_{2}
\le
\max\bigl\{\|\mathbf{v}_i-\mathbf{o}\|_{2},\;\|\mathbf{v}_{i+1}-\mathbf{o}\|_{2}\bigr\}. \label{eq:pmcm_range}
\end{equation}
In Figure~\ref{fig:method:window_points_filter}, $\mathbf{p}_j'$ violates Eq.~\eqref{eq:pmcm_range} for the window $\mathbf{v}_i'\mathbf{v}_{i+1}'$. The filter may be applied to a preprocessed cloud or to the ordered raw packet stream. Points that remain inside the corridor are passed to clustering, tracking, and then to \ac{gdeta}.

\subsection{Gated Dual-Evidence Threat Assessment}\label{subsec4.2}

\ac{gdeta} decides, from objects already inside the monitored motion corridor, whether the ego vehicle must command an emergency brake. The rule is deterministic and intended for embedded execution. Geometric occupancy and temporal urgency are scored independently; neither score is an input to the other. \textit{Hard gates} let either source command immediate braking. \textit{Soft scores} in $[0,1]$ are fused only if neither hard gate is active. The procedure consists of: (i)~a relevance gate; (ii)~geometric evidence scoring; (iii)~a geometric emergency gate; (iv)~temporal evidence scoring based on braking-oriented criticality measures and predicted corridor occupancy at arrival; (v)~an \ac{lfb} gate; (vi)~multi-object soft arbitration; and (vii)~the final brake decision.

\paragraph{Coordinate frame and inputs.}
Threat assessment is performed in the corridor-aligned frame of Sect.~\ref{subsec4.1}, with longitudinal arc-length coordinate $s$ along the ego reference path and lateral offset $d$ (Figure~\ref{fig:method:frenet_coordinate}). Let $\mathcal{O}$ be the set of tracked objects that already lie inside the monitored corridor and are supplied to \ac{gdeta}. For each object $i\in\mathcal{O}$, the upstream tracking module supplies the occupied footprint $B_i$ in the corridor frame, the nearest clearance $s_i^{\mathrm{near}}$, the stable-track flag $F_i$, the lateral center $d_i$, the longitudinal speed $v_{s,i}$, the lateral speed $v_{d,i}$, and the measured object width $w_i$. Envelope-occupancy persistence $n_i^{\mathrm{haz}}$ and soft-condition persistence $n_i^{\mathrm{soft}}$ are maintained by \ac{gdeta} from its own previous output on the same track identifier, matching the consecutive-count pattern of the \ac{lfb} gate. The \ac{ttd} used later is computed inside \ac{gdeta} from this body state.

\paragraph{Normalization.}
All soft scores are clipped to the unit interval by the saturation operator
\begin{equation}
\operatorname{sat}(x)=\min\bigl(\max(x,0),1\bigr),\label{eq:glra_sat}
\end{equation}
where $x\in\mathbb{R}$ is an arbitrary scalar. A linear map onto $[0,1]$ uses two thresholds: $x_0$ is the onset, at which the score leaves zero, and $x_1$ is the saturation, at which the score reaches one. If danger increases with $x$, then $x_0<x_1$ and
\begin{equation}
R=\operatorname{sat}\bigl((x-x_0)/(x_1-x_0)\bigr).\label{eq:gdeta_inc}
\end{equation}
If danger increases as $x$ decreases, then $x_0>x_1$ and
\begin{equation}
R=\operatorname{sat}\bigl((x_0-x)/(x_0-x_1)\bigr).\label{eq:gdeta_dec}
\end{equation}
Hence $R=0$ on the safe side of $x_0$, $R=1$ on the critical side of $x_1$, and $R$ interpolates linearly in between.

\paragraph{Relevance gate.}
From the input set $\mathcal{O}$, the relevance gate retains the tracks that enter threat assessment. Let $\gamma_i\in\{0,1\}$ be the ground-connection flag of object~$i$ and $\delta_i=\max(\ell_i^{\mathrm{meas}},w_i^{\mathrm{meas}})$ its observed spatial extent, the larger of the measured longitudinal and lateral footprint sizes. Two class-conditional minima are used, with $\delta_{\min}^{\mathrm{f}}>\delta_{\min}^{\mathrm{g}}>0$: $\delta_{\min}^{\mathrm{g}}$ for ground-connected objects and $\delta_{\min}^{\mathrm{f}}$ for floating objects. The relevant object set $\mathcal{O}_{\mathrm{rel}}$ collects those members of $\mathcal{O}$ that meet the matching size floor,
\begin{equation}
\mathcal{O}_{\mathrm{rel}}
=
\bigl\{
i\in\mathcal{O}:
(\gamma_i \land \delta_i\ge \delta_{\min}^{\mathrm{g}})
\;\lor\;
(\neg \gamma_i \land \delta_i\ge \delta_{\min}^{\mathrm{f}})
\bigr\}.\label{eq:gdeta_orel}
\end{equation}
Subsequent scoring and gates operate on $\mathcal{O}_{\mathrm{rel}}$.

\paragraph{Geometric evidence.}
For each relevant object $i\in\mathcal{O}_{\mathrm{rel}}$, a proximity score and a blocking score are defined by
\begin{align}
R_{\mathrm{prox},i}
&=
\operatorname{sat}\!\left(\frac{s_{\mathrm{act}}-s_i^{\mathrm{near}}}{\Delta s}\right),\label{eq:glra_prox}\\
R_{\mathrm{block},i}
&=
\operatorname{sat}\!\left(\frac{w_i^{\mathrm{ov}}}{w_{\mathrm{sat}}}\right),\label{eq:glra_block}
\end{align}
where $s_i^{\mathrm{near}}$ is the nearest longitudinal clearance from the ego front reference to object~$i$ along the corridor, $s_{\mathrm{act}}$ is the longitudinal activation distance at which geometric concern begins, $\Delta s>0$ is the longitudinal normalization span, $w_i^{\mathrm{ov}}$ is the lateral overlap width of object~$i$ with the monitoring corridor, and $w_{\mathrm{sat}}>0$ is the overlap width at which the blocking score saturates to one. The geometric evidence score is then
\begin{equation}
R_{\mathrm{geo},i}=R_{\mathrm{prox},i}\,R_{\mathrm{block},i}\in[0,1].\label{eq:glra_geo}
\end{equation}

\paragraph{Geometric emergency gate.}
The emergency envelope is the monitored corridor of Sect.~\ref{subsec4.1}, restricted to the braking-limited longitudinal interval $[0,s_{\mathrm{emg}}]$:
\begin{equation}
E_{\mathrm{emg}}
=
\Bigl\{
(s,d):\;
0\le s\le s_{\mathrm{emg}},\;
d_R(s)\le d\le d_L(s)
\Bigr\},\label{eq:glra_emg}
\end{equation}
with
\begin{equation}
s_{\mathrm{emg}}
=
v_{\mathrm{ego}}\,\tau_{\mathrm{sys}}
+
\frac{v_{\mathrm{ego}}^{2}}{2a_{\mathrm{avail}}}
+
s_{\mathrm{buf}}\label{eq:glra_semg}
\end{equation}
and with $d_L(s)$, $d_R(s)$ the left and right corridor bounds at arc length $s$. Here $v_{\mathrm{ego}}$ is the ego longitudinal speed, $\tau_{\mathrm{sys}}$ the command-path time reserve used in Eq.~\eqref{eq:glra_semg}, $a_{\mathrm{avail}}>0$ the available emergency deceleration, and $s_{\mathrm{buf}}$ a longitudinal safety buffer. 

Envelope occupancy $H_i\in\{0,1\}$ is one when the tracked occupied footprint $B_i$ intersects $E_{\mathrm{emg}}$ and the visible width meets a minimum,
\begin{equation}
H_i
=
\bigl(B_i\cap E_{\mathrm{emg}}\neq\emptyset\bigr)
\;\land\;
\bigl(w_i^{\mathrm{ov}}\ge w_{\min}\bigr),\label{eq:gdeta_h}
\end{equation}

Envelope-occupancy confirmation is the binary flag $C_i^{\mathrm{haz}}\in\{0,1\}$. \ac{gdeta} keeps a consecutive envelope-occupancy count $n_i^{\mathrm{haz}}$ on the same track identifier: the count increments while $H_i=1$ and resets otherwise, including when geometry is invalid. The flag is one when that count reaches the confirmation length $N_C^{\mathrm{geo}}$,
\begin{equation}
C_i^{\mathrm{haz}}
=
\mathbf{1}\bigl\{n_i^{\mathrm{haz}}\ge N_C^{\mathrm{geo}}\bigr\}.\label{eq:gdeta_chaz}
\end{equation}

Immediate geometric braking is reserved for envelope occupancy that is already both wide and close,
\begin{equation}
Q_i^{\mathrm{strong}}
=
\bigl(w_i^{\mathrm{ov}}\ge w_{\mathrm{block}}\bigr)
\;\land\;
\bigl(s_i^{\mathrm{near}}\le s_{\mathrm{imm}}\bigr),\label{eq:gdeta_qstrong}
\end{equation}
with $0<w_{\min}<w_{\mathrm{block}}$, and $s_{\mathrm{imm}}>0$ a near-contact gap. A confirmed tracker lock suppresses the geometric hard gate. Immediate geometric braking is commanded when
\begin{equation}
G_i^{\mathrm{geo}}
=
\neg F_i
\;\land\;
H_i
\;\land\;
\bigl(C_i^{\mathrm{haz}}\lor Q_i^{\mathrm{strong}}\bigr),\label{eq:glra_geo_cond}
\end{equation}
where $F_i\in\{0,1\}$ is the stable-track flag. 

\paragraph{Temporal evidence.}
Temporal evidence is computed independently of the geometric evidence. Longitudinal closing uses the constant-acceleration relative-gap model
\begin{equation}
g_i(t)=s_{\mathrm{gap},i}-v_{\mathrm{rel},i}\,t-\tfrac{1}{2}a_{\mathrm{rel},i}\,t^{2},\label{eq:glra_gap}
\end{equation}
where $g_i(t)$ is the predicted longitudinal clearance at time~$t\ge 0$, $s_{\mathrm{gap},i}$ is the current longitudinal gap to object~$i$, $v_{\mathrm{rel},i}=v_{\mathrm{ego}}-v_{\mathrm{obj},i}$ is the relative closing speed (positive when closing), and $a_{\mathrm{rel},i}=a_{\mathrm{ego}}-a_{\mathrm{obj},i}$ is the relative longitudinal acceleration. The \ac{ttc} is the smallest positive root of $g_i(t)=0$. From these kinematics, the required braking magnitude $a_{\mathrm{req},i}$ yields the \ac{btn}
\begin{equation}
\acmath{btn}_i=\frac{a_{\mathrm{req},i}}{a_{\mathrm{avail}}},\label{eq:glra_btn}
\end{equation}
where $a_{\mathrm{req},i}\ge 0$ is the minimum ego braking deceleration needed immediately to avoid a predicted collision. The \ac{ttb} is the latest admissible delay before applying $a_{\mathrm{avail}}$ that still avoids collision. If no collision is predicted, $\acmath{ttc}_i=\acmath{ttb}_i=\infty$ and $\acmath{btn}_i=0$. Normalized temporal sub-scores $R_{\acmath{ttb},i},R_{\acmath{btn},i}\in[0,1]$ are obtained by mapping $\acmath{ttb}_i$ with Eq.~\eqref{eq:gdeta_dec} and $\acmath{btn}_i$ with Eq.~\eqref{eq:gdeta_inc} between warning ($x_0$) and critical ($x_1$) thresholds. Those longitudinal metrics measure closing urgency. 

The front-arrival time $t_i^{\mathrm{arr}}$ is the first instant at which Eq.~\eqref{eq:glra_gap} reaches the object's current longitudinal position. Three outcomes are possible:
\begin{equation}
t_i^{\mathrm{arr}}
=
\begin{cases}
0 & \text{if }s_{\mathrm{gap},i}\le 0,\\
TTC & \text{if such a }t\text{ exists},\\
\text{undefined} & \text{if }s_{\mathrm{gap},i}>0\text{ and no such }t\text{ exists.}
\end{cases}\label{eq:gdeta_tarr}
\end{equation}
The corridor width at position $s$ is the span of the bounds already used in Eq.~\eqref{eq:glra_emg},
\begin{equation}
w_{\mathrm{c}}(s)=d_L(s)-d_R(s).\label{eq:gdeta_wc}
\end{equation}
The lateral exit offset $r_i=\tfrac12(w_{\mathrm{c}}+w_i)$ is the distance the object center must travel from the corridor centerline to leave the corridor. Arrival occupation $R_{\mathrm{occ},i}$ is one if the object center is still inside the corridor, by at least the leftover-stay margin, when the front arrives. 
\begin{equation}
R_{\mathrm{occ},i}
=
\mathbf{1}\bigl\{\bigl|d_i+v_{d,i}\,t_i^{\mathrm{arr}}\bigr|\le r_i-\lvert v_{d,i}\rvert\tau_{\mathrm{occ}}\bigr\}.\label{eq:glra_occ}
\end{equation}
The margin $\tau_{\mathrm{occ}}$ is the minimum leftover stay counted as occupied. If $t_i^{\mathrm{arr}}$ is undefined, $R_{\mathrm{occ},i}=0$. Leftover stay is the time the object remains in the corridor after front arrival. The object leaves the corridor at 
\begin{equation}
\acmath{ttd}_i
=
\sup\bigl\{t\ge 0:\bigl|d_i+v_{d,i}\,t\bigr|\le r_i\bigr\},\label{eq:gdeta_ttd}
\end{equation}
with $\acmath{ttd}_i=\infty$ if the object remains in the band for all future time. Then $\tau_{\mathrm{stay},i}=\acmath{ttd}_i-t_i^{\mathrm{arr}}$ when the body occupies the raw band at $t_i^{\mathrm{arr}}$. The fused temporal evidence score is

\begin{equation}
R_{\mathrm{temp},i}
=
R_{\mathrm{occ},i}\,
\Bigl(
1-(1-R_{\acmath{ttb},i})(1-R_{\acmath{btn},i})
\Bigr)\in[0,1].\label{eq:glra_temp}
\end{equation}

\paragraph{\ac{lfb} gate.}
The flag $V_i$ admits the \ac{lfb} gate only when the object is still in the corridor at front arrival and remains there for at least the leftover-stay threshold $T_m>\tau_{\mathrm{occ}}$:
\begin{equation}
V_i=\mathbf{1}\bigl\{R_{\mathrm{occ},i}=1 \land \tau_{\mathrm{stay},i}\ge T_m\bigr\}.\label{eq:gdeta_vi}
\end{equation}
A scan is urgent only when $V_i=1$ and either remaining brake delay or required deceleration is already critical,
\begin{equation}
\mathcal{R}_i^{\mathrm{LFB,now}}
=
V_i
\land
\Bigl(
\bigl(\acmath{ttb}_i\le\tau_{\mathrm{act}}+\tau_{\mathrm{buf}}\bigr)
\;\lor\;
\bigl(\acmath{btn}_i\ge \theta_{btn}\bigr)
\Bigr),\label{eq:glra_lfb_cond}
\end{equation}
where $\tau_{\mathrm{act}}$ is the brake actuation delay, $\tau_{\mathrm{buf}}$ is a temporal safety buffer, and $\theta_{btn}=\mathrm{BTN}_{\mathrm{crit}}=1$ is the saturation threshold of required versus available deceleration. Either longitudinal clause is sufficient; both are gated by $V_i$. The consecutive count $n_i^{\mathrm{risk}}$ increments while $\mathcal{R}_i^{\mathrm{LFB,now}}$ holds for the same identifier and resets otherwise. Immediate \ac{lfb} braking is commanded when
\begin{equation}
G_i^{\acmath{lfb}}
=
\mathbf{1}\bigl\{n_i^{\mathrm{risk}}\ge N_C^{\acmath{lfb}}\bigr\},\label{eq:gdeta_lfb_gate}
\end{equation}
so the gate confirms over $N_C^{\acmath{lfb}}$ scans. 

\paragraph{Soft arbitration and final decision.}
For objects on which neither hard gate has fired, a per-object \ac{ri} is formed by soft-OR fusion,
\begin{equation}
\acmath{ri}_i=1-(1-R_{\mathrm{geo},i})(1-R_{\mathrm{temp},i}),\label{eq:glra_ri}
\end{equation}
and the scene threat index is taken as the dominant-object value
${\acmath{ri}}_{\mathrm{scene}}=\max_{i}\acmath{ri}_i$.
The instantaneous soft condition on object~$i$ requires arrival occupation, a positive temporal score, and an index at or above the threshold,
\begin{equation}
S_i^{\mathrm{now}}
=
\bigl(R_{\mathrm{occ},i}=1\bigr)
\land
\bigl(R_{\mathrm{temp},i}>0\bigr)
\land
\bigl(\acmath{ri}_i\ge\theta_{\acmath{ri}}\bigr),\label{eq:gdeta_snow}
\end{equation}
where $\theta_{\acmath{ri}}\in(0,1]$ is a soft braking threshold. The consecutive count $n_i^{\mathrm{soft}}$ increments while $S_i^{\mathrm{now}}$ holds for the same identifier and resets otherwise. Let $i^\star$ be the dominant object in the residual soft set. The ego brake command is issued when
\begin{equation}
\mathrm{Brake}
\;\Leftarrow\;
\Bigl(\exists i:\,G_i^{\mathrm{geo}}=1\Bigr)
\;\lor\;
\Bigl(\exists i:\,G_i^{\acmath{lfb}}=1\Bigr)
\;\lor\;
\bigl(n_{i^\star}^{\mathrm{soft}}\ge N_C^{\mathrm{ri}}\bigr).\label{eq:glra_brake}
\end{equation}

\section{Experimental Evaluation and Validation}\label{sec:eval}
\subsection{Evaluation of \texorpdfstring{\acs{pmcm}}{PMCM}}\label{subsec:pmcm_eval}

Section~\ref{subsec4.1} constructs a path-conforming swept-body envelope and retains those \ac{lidar} returns that fall inside it. That construction is the geometric precondition for \ac{gdeta}, which consumes tracked objects already inside the monitored set. The experiments below ask whether, relative to simpler masks, the \ac{pmcm} envelope covers the predicted body sweep tightly enough that a static obstacle on the corridor edge is detected in time. 

\subsubsection{Controlled Collision Simulation}

\begin{figure}[th]
  \centering
  \includegraphics[width=\textwidth]{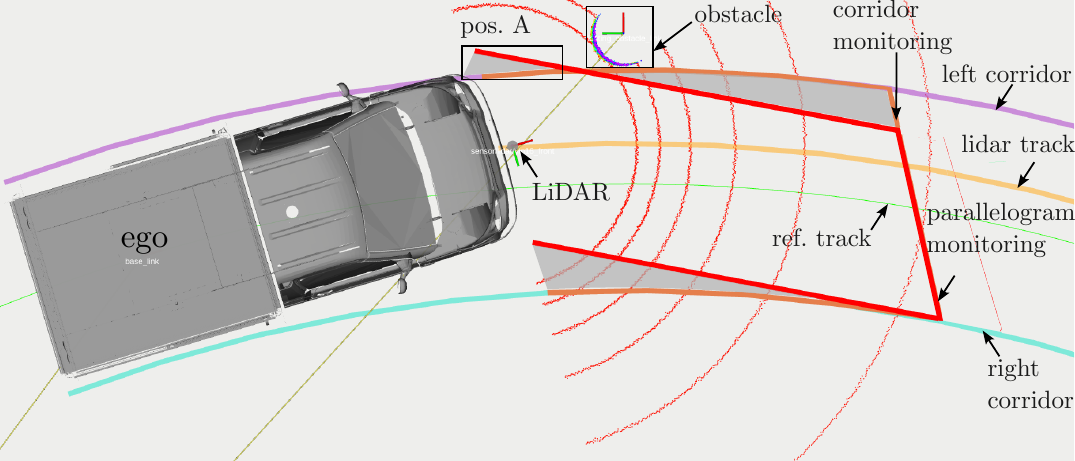}
  \caption{Bird's-eye view of a simulated corridor-edge collision case. The parallelogram mask, the \ac{pmcm} corridor, and the first-detection location \textit{pos.\ A} are indicated~\cite{pan2026enhancing}.}\label{fig:eval:pmcm_sim}
\end{figure}

\paragraph{Experimental setup.}
The study is performed in Gazebo with a sensor model of a Velodyne VLP-16 that matches the experimental vehicle. A static cylinder of radius $0.35\,\mathrm{m}$ and height $0.6\,\mathrm{m}$ is placed on the outer edge of the driving corridor. In each trial the ego vehicle holds a constant speed on either a circular path or a straight road. Speeds range from $1\,\mathrm{km/h}$ to $30\,\mathrm{km/h}$ in steps of $2\,\mathrm{km/h}$. The path radii are $\{6,8,10,12,14,16,18,20,22,24,26,\infty\}\,\mathrm{m}$, where an infinite radius denotes the straight-road case. The simulated vehicle uses the same body dimensions as the experimental platform. For every trial the first detected obstacle position is stored, and it is recorded whether a subsequent emergency stop still avoids contact. All designed encounters are collision cases, so the comparison uses the \ac{fnr}, the \ac{tpr}, and the residual \ac{cr} after a commanded stop. Figure~\ref{fig:eval:pmcm_sim} shows one such encounter, together with the parallelogram mask, the corridor mask, and the first-detection location marked \textit{pos.\ A}.

\paragraph{Reference filters.}
Two alternative masks are evaluated on the same trials. The first is a parallelogram filter that adapts a quadrilateral keep-out region to the instantaneous driving state. The ratio of parallelogram area to \ac{pmcm} corridor area was computed for each radius. The ratio increases toward unity as the path straightens, and it quantifies how much corridor area the parallelogram loses on tight turns. The second baseline, two-point corridor filtering, traces only the two rear vehicle corners along the predicted path and treats the resulting strip as the monitor. It has a corridor-like outline, yet it does not apply the sliding-extrema envelope of Eq.~\eqref{eq:pmcm_extrema}.

\begin{table}[t]
\centering
\caption{Comparison of \ac{fnr}, \ac{tpr}, and residual \ac{cr} for three monitoring masks on the same controlled collision series. Adapted from~\cite{pan2026enhancing}.}
\label{tab:eval:pmcm_filters}
\begin{tabular}{@{}lccc@{}}
\toprule
Measure & Parallelogram & Two-point corridor & \acmath{pmcm} \\
\midrule
\ac{fnr}  & $0\%$    & $57.74\%$ & $\mathbf{0}\%$ \\
\ac{tpr}  & $100\%$  & $42.26\%$ & $\mathbf{100}\%$ \\
\ac{cr}   & $36.1\%$ & $1.03\%$  & $\mathbf{0}\%$ \\
\botrule
\end{tabular}
\end{table}
\paragraph{Detection timing and collision outcome.}

Table~\ref{tab:eval:pmcm_filters} summarizes the three masks. Both the parallelogram filter and \ac{pmcm} attain a \ac{fnr} of $0\%$ and a \ac{tpr} of $100\%$: the cylinder is eventually seen in every trial. Their residual \ac{cr} values differ. The parallelogram mask leaves a collision in $36.1\%$ of the trials, whereas \ac{pmcm} leaves none. The recorded first-detection poses show why. Under the parallelogram mask the cylinder often becomes visible only at \textit{pos.\ A} in Figure~\ref{fig:eval:pmcm_sim}, which already lies inside the braking distance $d_b$. A stop commanded at that instant cannot prevent impact. As curvature increases, the parallelogram occupies a shrinking fraction of the true swept corridor, so an obstacle on the outer edge enters the mask too late even though the object is eventually detected. On large radii the same ratio approaches one, and the parallelogram becomes an acceptable stand-in for the path-conforming envelope.

The two-point strip attains a residual \ac{cr} of $1.03\%$, close to that of \ac{pmcm}, but it misses the cylinder in $57.74\%$ of the trials. Most of those misses occur at radii of $16\,\mathrm{m}$ or less. Two rear-corner generators cannot follow the outer body sweep on a tight turn, so a corridor-edge obstacle may lie outside the strip for the entire approach. That is a different failure mode from the parallelogram case: the two-point mask often never sees the object, whereas the parallelogram sees it too late. \ac{pmcm} avoids both modes because the sliding-extrema envelope of Sect.~\ref{subsec4.1} is built from the vehicle geometry itself.

\subsubsection{Recorded Campus Driving}

\begin{figure}[t]
    \centering
    \includegraphics[width=\textwidth]{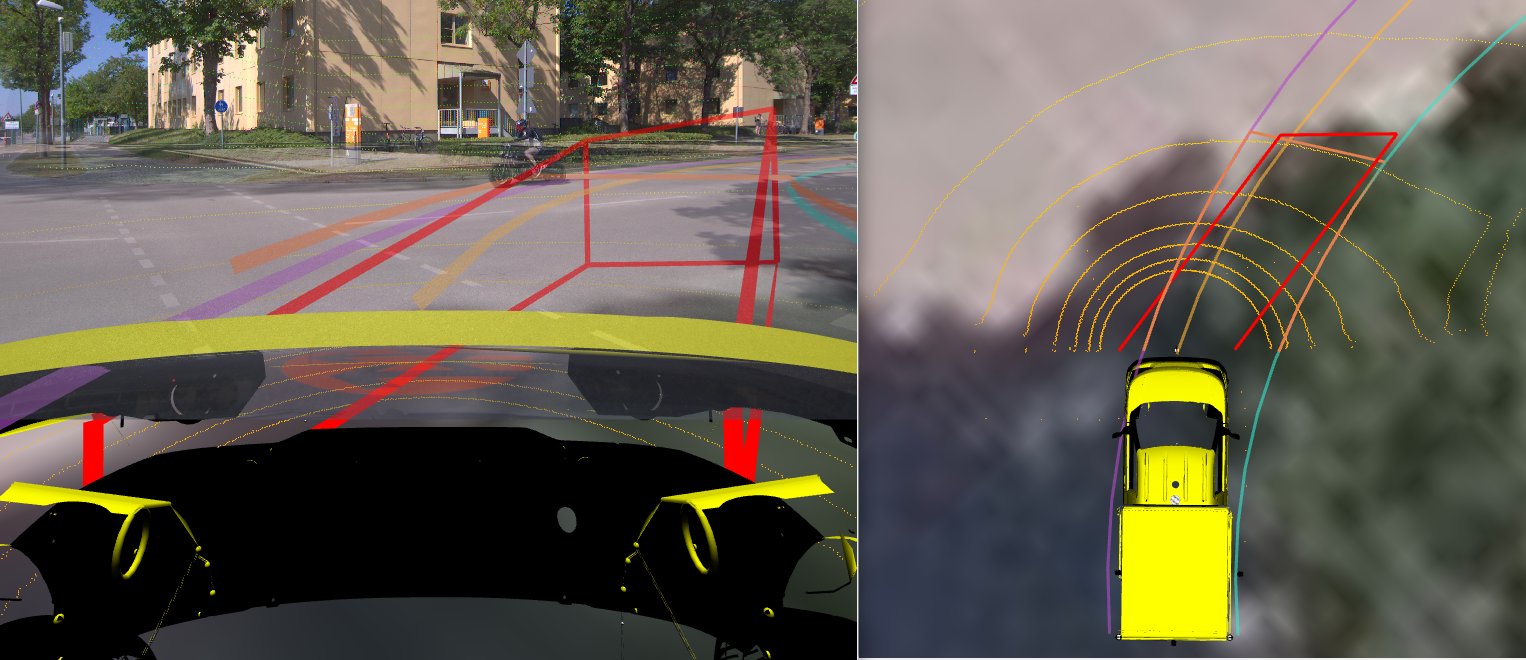}
    \caption{Online \ac{pmcm} corridor on recorded campus driving, shown in a camera view and a bird's-eye overlay~\cite{pan2026enhancing}.}\label{fig:eval:pmcm_campus}
\end{figure}

The same envelope is next applied to recorded campus runs, which include urban-like turns, sensor noise, ego-motion fluctuation, and localization drift that the Gazebo series does not reproduce. Figure~\ref{fig:eval:pmcm_campus} shows one snippet in a camera view and a corresponding bird's-eye overlay. The left and right corridor polylines update continuously along the predicted path, remain aligned with the vehicle body sweep, and follow changes in turning radius without switching mask type. The recording is used to check that the online boundary stays aligned under those disturbances. 

\subsection{Evaluation of \texorpdfstring{\acs{gdeta}}{GDETA}}\label{subsec:glra_eval}

\ac{gdeta} consumes tracked objects that are already inside the monitored set~$\mathcal{O}$. This subsection tests the decision rule of Sect.~\ref{subsec4.2} on object-level states already expressed in the corridor frame. Two cases are studied. An established lead has a stable track, so the geometric hard gate is suppressed and the trial tests temporal and soft braking. A transverse crossing first appears at the corridor boundary; this is the sudden corridor entry the monitor is meant to cover, and the trial tests the geometric gate and its later handover to temporal assessment.

\subsubsection{Evaluation Design}
The decision logic is evaluated in a Monte Carlo simulation. Each trial is split into three parallel worlds.

\paragraph{Ground-truth world.}
This world propagates the ego and one object at $100\,\mathrm{Hz}$. In the lead case both bodies use constant acceleration along the path: the lead cruises, then brakes after a sampled time. In the crossing case the target holds along-path speed at zero and moves through the corridor at constant lateral speed. Two noise-free rollouts supply labels. The no-brake rollout keeps constant ego speed and marks the trial dangerous if the bodies meet. The oracle is a second rollout in which the ego applies the same delayed first-order brake (a $0.2\,\mathrm{s}$ dead time and a $0.15\,\mathrm{s}$ rise toward $7.5\,\mathrm{m/s^2}$) at the first instant a causal reaction is defined. For lead braking that instant is a physics event: the sampled time when the lead starts to brake. For crossing, the target is invisible until its body overlaps the corridor, and the monitor only looks at $10\,\mathrm{Hz}$ scans, so the oracle latches on the first scan at which that overlap exists. A remaining oracle collision is labeled unavoidable and is excluded from the avoidable-collision denominator.

\paragraph{\ac{gdeta} input world.}
This world converts the current ground-truth state into corridor-aligned object metadata. %
% : gap, speeds, accelerations, lateral centre, visible width, and the stable-track flag $F_i$. 
Independent Gaussian noise is added only here, on those decision inputs, to emulate imperfect tracking. Both campaigns use standard deviations of $0.5\,\mathrm{m}$, $0.5\,\mathrm{m/s}$, and $0.5\,\mathrm{m/s^2}$ on gap, longitudinal speed, and longitudinal acceleration. Crossing also uses $0.1\,\mathrm{m/s}$ on lateral speed. Its configuration records a $0.05\,\mathrm{m}$ lateral-position deviation. For the lead case, $F_i=1$ from the first scan. For the crossing case, motion state is invalid on the first two visible scans, and $F_i$ stays $0$ for three visible scans and then switches to $1$ to indicate a valid track. At every $10\,\mathrm{Hz}$ scan the method returns a brake command together with the geometric and temporal scores and the hard-gate flags of Sect.~\ref{subsec4.2}.

\paragraph{Evaluation world.}
This world closes the loop around \ac{gdeta} on the same ground-truth plant. After the latch, ego deceleration follows the actuator model above. The run stops at impact or mutual standstill. The closed-loop outcome is judged against the two ground-truth counterfactuals: the no-brake rollout defines whether the sample is dangerous, and the oracle flags collisions that remain after that earliest command.

\subsubsection{Monte Carlo Sampling}

The campaign covers the two cases listed in Table~\ref{tab:eval:gdeta_protocol}. Continuous quantities are sampled independently from uniform distributions. The speed range of $3$--$30\,\mathrm{km/h}$ matches the intended near-field operation, while the lead-vehicle deceleration range of $[3.0,7.5]\,\mathrm{m/s^2}$ covers moderate through emergency braking. The \ac{gdeta} settings are fixed: $N_C^{\mathrm{geo}}=2$, $N_C^{\acmath{lfb}}=3$, $N_C^{\mathrm{ri}}=2$, $\theta_{\acmath{ri}}=0.89$, $a_{\mathrm{avail}}=7.5\,\mathrm{m/s^2}$, $T_{\mathrm{warn}}=0.6\,\mathrm{s}$, $T_{\mathrm{crit}}=0.08\,\mathrm{s}$, $\mathrm{BTN}_{\mathrm{warn}}=0.6$, and $\mathrm{BTN}_{\mathrm{crit}}=1.0$.

\begin{table*}[t]
\caption{Object-level Monte Carlo scene design. Continuous quantities are sampled independently from uniform distributions.}
\label{tab:eval:gdeta_protocol}
\begin{tabular}{@{}p{0.21\textwidth}p{0.36\textwidth}p{0.36\textwidth}@{}}
\toprule
Property & Established lead braking & Transverse crossing \\
\midrule
Trials / seed & $200{,}000$ / $10$ & $200{,}000$ / $10$ \\
Horizon & $60\,\mathrm{s}$ & $12\,\mathrm{s}$ \\
Ego speed & $3$--$30\,\mathrm{km/h}$ & $3$--$30\,\mathrm{km/h}$ \\
Target body & $3.5\times2.0\,\mathrm{m}$ vehicle & $0.5\times0.4\,\mathrm{m}$ target \\
Target motion & $0$--$30\,\mathrm{km/h}$;  $3$--$7.5\,\mathrm{m/s^2}$ braking & $v_s=0$; $\lvert v_d\rvert=0.5$--$3\,\mathrm{m/s}$ \\
Encounter timing & gap $0.5$--$15\,\mathrm{m}$; brake onset $0$--$5\,\mathrm{s}$ & centerline-arrival offset $-3$--$3\,\mathrm{s}$ \\
Flag $F_i$ & $1$ from the first scan & $0$ for three visible scans, then $1$ \\
\botrule
\end{tabular}
\end{table*}

In the lead case, the leading vehicle remains centered with zero lateral velocity, so $R_{\mathrm{occ},i}=1$, $\tau_{\mathrm{stay},i}=\infty$, and $V_i=1$ throughout a typical trial. The fixed $F_i=1$ suppresses $G_i^{\mathrm{geo}}$, and any command must come from the \ac{lfb} gate or soft arbitration. In the crossing case the target starts just outside the corridor, left or right with equal probability, and has no along-path speed. Its initial gap is computed from a sampled arrival offset in $[-3,3]\,\mathrm{s}$: offset zero means the ego front reaches the target's fixed path position at the same instant the target reaches the corridor centerline. The geometric gate may fire while $F_i=0$ and motion is still invalid; once $F_i=1$, Eq.~\eqref{eq:glra_geo_cond} suppresses that gate and later scans use the temporal and soft paths.

\subsubsection{Results and Discussion}

Table~\ref{tab:eval:gdeta_results} reports the simulation results. Brake-command recall is computed among real collisions (\ac{tp} and \ac{fn}) that the causal oracle can avoid. A false positive is a command in a no-brake non-collision trial.

\begin{table*}[t]
\caption{\ac{gdeta} Monte Carlo results. }
\label{tab:eval:gdeta_results}
\begin{tabular}{@{}p{0.43\textwidth}p{0.25\textwidth}p{0.25\textwidth}@{}}
\toprule
Metrics & Lead braking & Transverse crossing \\
\midrule
Trials & $200{,}000$ & $200{,}000$ \\
Unavoidable cases  & $59{,}690$  & $93{,}840$ \\
\ac{fnr} & $0.05\%$ & $24.20\%$ \\
brake-command recall & $99.95\%$ & $75.80\%$ \\
precision & $99.99\%$ & $97.05\%$ \\
\ac{fpr} & $4/18$ ($22.22\%$) & $799/71{,}497$ ($1.12\%$) \\
\botrule
\end{tabular}
\end{table*}

In the lead-braking campaign, \ac{gdeta} identifies nearly all avoidable longitudinal conflicts, missing $75$ of $140{,}292$ cases. Across all lead trials, including cases classified by the oracle as unavoidable, the first command comes from soft arbitration in $199{,}484$ cases and from the \ac{lfb} gate in $406$ cases. With the stable-track flag fixed at $F_i=1$, none comes from the geometric gate. The campaign contains only $18$ no-brake non-collisions, so the observed $4/18$ \ac{fpr} is a thin sample of nuisance braking. The missed cases are low-speed approaches to a stopped lead vehicle, for which both \ac{ttb} and \ac{btn} remain below their activation thresholds until contact. The temporal urgency maps therefore miss that low-speed contact condition.

In the transverse-crossing campaign the two hard gates and soft arbitration all fire. The method commands braking in $26{,}276$ of the $34{,}663$ avoidable collision cases. Among the $39{,}475$ crossing commands, the first command is geometric-only in $10{,}148$ cases, temporal-only in $7{,}017$, and soft-only in $22{,}310$. Newly visible targets can therefore reach the geometric gate before $F_i$ suppresses it, and later scans can use the temporal or soft rule. Of the $71{,}497$ no-brake non-collision trials, $799$ receive a command, so the \ac{fpr} is $1.12\%$. The avoidable-collision \ac{fnr} is $24.20\%$. An emergency stop on a non-collision interrupts the planned motion, so the consecutive-scan gates and $\theta_{\acmath{ri}}$ keep the false-alarm rate low. The resulting \ac{fn}/\ac{fp} operating point is that of a redundant last-resort brake: a crossing that remains visible for several scans is a case the \ac{hlas} is expected to treat first, and \ac{nefimmore} is reserved for a sudden corridor intrusion that the roof suite cannot process in time. No intervention induces a collision in either campaign.

After three visible scans $F_i$ becomes $1$ and the geometric gate closes, so later braking needs consecutive urgent frames from the \ac{lfb} gate or from soft arbitration. A $0.4\,\mathrm{m}$ pedestrian reaches that urgency only when the gap is already small. When a command does appear, it is on average $0.35\,\mathrm{s}$ before the no-brake impact, with a mean \ac{ttb} of $0.048\,\mathrm{s}$.

\section{Limitations}\label{sec5}

\ac{nefimmore} is a redundant last-resort monitor of the body sweep predicted from the \ac{hlas} path. Objects outside the motion corridor are excluded from its decision scope. Restricting the monitored region is a trade-off between computational demand and coverage. The method responds to hazards that enter the corridor, and its decisions rely on a single 3D \ac{lidar}. Consecutive-scan confirmation in \ac{gdeta} is set to keep nuisance emergency stops low; the \ac{hlas} remains the first actor on crossings it can observe in time.

The geometric evidence score in \ac{gdeta} varies linearly with the front gap. A nonlinear mapping may be more appropriate when the decision should become more sensitive as the gap approaches zero.

The simulation considers one object at a time: either a longitudinally braking lead vehicle or a lateral target moving at constant speed. The longitudinal model assumes constant relative acceleration, while the crossing model assumes constant lateral speed. Multi-object interactions and lane-entry scenarios are left to future work.

\section{Conclusion}\label{sec6}

This article combines \acf{pmcm} and \acf{gdeta} in one \acf{nefimmore} system. \ac{pmcm} constructs a curvature-aware monitoring boundary from the path supplied by the \acf{hlas} and the vehicle geometry, and uses that boundary to filter \ac{lidar} returns. \ac{gdeta} then decides whether the ego vehicle must command an emergency brake for tracked objects in the same corridor. Geometric occupancy and temporal urgency stay separate: either hard gate can issue a brake command, and soft arbitration combines the two scores when both gates remain inactive.

\ac{pmcm} is evaluated by placing an obstacle at the outer corridor edge over a range of speeds and path radii. The proposed boundary detects every tested obstacle; the comparison masks either see it too late or miss it. Object-level Monte Carlo trials then test \ac{gdeta} in established lead-braking and transverse-crossing cases. Lead-vehicle brake-command recall is $99.95\%$. In the crossing case all three decision mechanisms fire, and the false-alarm rate among non-collisions is $1.12\%$. The \ac{hlas} remains the first actor on crossings it can observe in time; \ac{nefimmore} is reserved for a sudden corridor intrusion.

\backmatter

\bmhead{Acknowledgements}

This research is funded by dtec.bw – Digitalization and Technology Research Center of the Bundeswehr, which we gratefully acknowledge. dtec.bw is funded by the European Union – NextGenerationEU. We also acknowledge financial support from the FORscience Research Fund of the Universität der Bundeswehr München. The authors thank Matthias Gerdts and Thomas Rottmann for valuable discussions, and the anonymous reviewers for their feedback.

\section*{Declarations}
\bmhead{Competing interests}
The authors declare that they have no competing financial or other interests that could influence the research or its outcomes.

\bmhead{Ethical Approval}
This research did not involve human participants, animals, or sensitive data, so ethical approval was not required.

\bmhead{Consent to Participate}
Not applicable. This study did not involve human participants.

\bmhead{Consent for Publication}
Not applicable. This study did not involve human participants.

\bigskip

\bibliography{sn-bibliography}% common bib file

@article{yang2022systematic,
  author  = {Yang, Lin and Yang, Ying and Wu, Guoyuan and Zhao, Xiangmo and Fang, Shan and Liao, Xianghong and Wang, Rui and Zhang, Meng},
  title   = {A Systematic Review of Autonomous Emergency Braking System: Impact Factor, Technology, and Performance Evaluation},
  journal = {Journal of Advanced Transportation},
  volume  = {2022},
  pages   = {1--13},
  year    = {2022},
  doi     = {10.1155/2022/1188089}
}

@article{kaempchen2009situation,
  author  = {Kaempchen, Nico and Schiele, Bernt and Dietmayer, Klaus},
  title   = {Situation Assessment of an Autonomous Emergency Brake for Arbitrary Vehicle-to-Vehicle Collision Scenarios},
  journal = {IEEE Transactions on Intelligent Transportation Systems},
  volume  = {10},
  number  = {4},
  pages   = {678--687},
  year    = {2009},
  doi     = {10.1109/TITS.2009.2026452}
}

@article{brannstrom2010model,
  author  = {Br{\"a}nnstr{\"o}m, Mattias and Coelingh, Erik and Sj{\"o}berg, Jonas},
  title   = {Model-Based Threat Assessment for Avoiding Arbitrary Vehicle Collisions},
  journal = {IEEE Transactions on Intelligent Transportation Systems},
  volume  = {11},
  number  = {3},
  pages   = {658--669},
  year    = {2010},
  doi     = {10.1109/TITS.2010.2048314}
}

@article{han2024adaptive,
  author  = {Han, S. and Choi, K. H. and Kim, K.-S.},
  title   = {Adaptive Safety Distance for Collision Avoidance Based on a Full Braking Activation Parameter},
  journal = {IEEE Transactions on Intelligent Transportation Systems},
  volume  = {25},
  number  = {9},
  pages   = {12657--12668},
  year    = {2024},
  doi     = {10.1109/TITS.2024.3386165}
}

@article{hanselaar2024safety,
  author  = {Hanselaar, C. A. J. and Silvas, E. and Terechko, A. and Heemels, W. P. M. H.},
  title   = {The Safety Shell: An Architecture to Handle Functional Insufficiencies in Automated Driving},
  journal = {IEEE Transactions on Intelligent Transportation Systems},
  volume  = {25},
  number  = {7},
  pages   = {7522--7540},
  year    = {2024},
  doi     = {10.1109/TITS.2024.3352829}
}

@article{pek2020using,
  author  = {Pek, Christian and Manzinger, Stefanie and Koschi, Markus and Althoff, Matthias},
  title   = {Using Online Verification to Prevent Autonomous Vehicles from Causing Accidents},
  journal = {Nature Machine Intelligence},
  volume  = {2},
  number  = {9},
  pages   = {518--528},
  year    = {2020},
  doi     = {10.1038/s42256-020-0225-y}
}

@article{pek2021fail,
  author  = {Pek, Christian and Althoff, Matthias},
  title   = {Fail-Safe Motion Planning for Online Verification of Autonomous Vehicles Using Convex Optimization},
  journal = {IEEE Transactions on Robotics},
  volume  = {37},
  number  = {3},
  pages   = {798--814},
  year    = {2021},
  doi     = {10.1109/TRO.2020.3036624}
}

@article{kumar2024autonomous,
  author  = {Kumar, V. and Subramanian, S. C. and Rajamani, R.},
  title   = {Autonomous Emergency Braking of a Heavy Road Vehicle Using a Low-Density Flash Lidar},
  journal = {IEEE Transactions on Vehicular Technology},
  volume  = {73},
  number  = {2},
  pages   = {1879--1889},
  year    = {2024},
  doi     = {10.1109/TVT.2023.3314534}
}

@article{pan2023collision,
  author  = {Pan, Junnan and Sotiriadis, Prodromos and Englberger, Ferdinand},
  title   = {Collision Avoidance of Autonomous Driving at Low Speed in the Near Field of Vehicle},
  journal = {International Journal of Emerging Engineering and Technology},
  volume  = {2},
  number  = {1},
  pages   = {57--62},
  year    = {2023},
  doi     = {10.57041/ijeet.v2i1.916}
}

@inproceedings{pan2024independent,
  author    = {Pan, Junnan and Sotiriadis, Prodromos and Englberger, Ferdinand},
  title     = {Independent Near-Field Monitoring: A Novel Approach to Improve Active Safety in Autonomous Vehicles},
  booktitle = {Proc. IEEE Intelligent Vehicles Symposium (IV)},
  address   = {Jeju Island, Korea},
  pages     = {730--736},
  year      = {2024},
  doi       = {10.1109/IV55156.2024.10588775}
}

@inproceedings{pan2025reliability,
  author    = {Pan, Junnan and Sotiriadis, Prodromos and Nenchev, Vladislav and Englberger, Ferdinand},
  title     = {Improving Functional Reliability of Near-Field Monitoring for Emergency Braking in Autonomous Vehicles},
  booktitle = {2025 IEEE 28th International Conference on Intelligent Transportation Systems (ITSC)},
  year      = {2025},
  doi       = {10.1109/ITSC60802.2025.11423625}
}

@inproceedings{pan2026enhancing,
  author    = {Pan, Junnan and Mansharamani, Mohak and Sotiriadis, Prodromos and Nenchev, Vladislav and Englberger, Ferdinand},
  title     = {Enhancing Functional Reliability of Autonomous Vehicle Safety Monitoring in Curves},
  booktitle = {Proc. 12th International Conference on Vehicle Technology and Intelligent Transport Systems (VEHITS)},
  year      = {2026},
  doi     = {10.5220/0014812400004030}
}

@inproceedings{gutjahr2014automatic,
  author    = {Gutjahr, Benjamin and Werling, Moritz},
  title     = {Automatic Collision Avoidance during Parking and Maneuvering --- An Optimal Control Approach},
  booktitle = {Proc. IEEE Intelligent Vehicles Symposium (IV)},
  address   = {Dearborn, MI, USA},
  pages     = {636--641},
  year      = {2014},
  doi       = {10.1109/IVS.2014.6856575}
}

@article{dahl2019collision,
  author  = {Dahl, Jonas and de Campos, Gabriel Rodrigues and Olsson, C. and Fredriksson, Jonas},
  title   = {Collision Avoidance: A Literature Review on Threat-Assessment Techniques},
  journal = {IEEE Transactions on Intelligent Vehicles},
  volume  = {4},
  number  = {1},
  pages   = {101--113},
  year    = {2019},
  doi     = {10.1109/TIV.2018.2886682}
}

@article{westhofen2023criticality,
  author  = {Westhofen, Lukas and Neurohr, Christian and Koopmann, Tjark and Butz, Martin and Sch{\"u}tt, Barbara and Utesch, Fabian and Neurohr, Birte and Gutenkunst, Christian and B{\"o}de, Eckard},
  title   = {Criticality Metrics for Automated Driving: A Review and Suitability Analysis of the State of the Art},
  journal = {Archives of Computational Methods in Engineering},
  volume  = {30},
  number  = {1},
  pages   = {1--35},
  year    = {2023},
  doi     = {10.1007/s11831-022-09788-7}
}

@article{jansson2008framework,
  author  = {Jansson, Jonas and Gustafsson, Fredrik},
  title   = {A Framework and Automotive Application of Collision Avoidance Decision Making},
  journal = {Automatica},
  volume  = {44},
  number  = {9},
  pages   = {2347--2351},
  year    = {2008},
  doi     = {10.1016/j.automatica.2008.01.016}
}

@article{gelso2017consistent,
  author  = {Gelso, Esteban R. and Sj{\"o}berg, Jonas},
  title   = {Consistent Threat Assessment in Rear-End Near-Crashes Using {BTN} and {TTB} Metrics, Road Information and Naturalistic Traffic Data},
  journal = {IEEE Intelligent Transportation Systems Magazine},
  volume  = {9},
  number  = {1},
  pages   = {74--89},
  year    = {2017},
  doi     = {10.1109/MITS.2016.2550012}
}

@article{falcone2011predictive,
  author  = {Falcone, Paolo and Ali, Mohammad and Sj{\"o}berg, Jonas},
  title   = {Predictive Threat Assessment via Reachability Analysis and Set Invariance Theory},
  journal = {IEEE Transactions on Intelligent Transportation Systems},
  volume  = {12},
  number  = {4},
  pages   = {1352--1361},
  year    = {2011},
  doi     = {10.1109/TITS.2011.2158210}
}

@inproceedings{sontges2018worst,
  author    = {S{\"o}ntges, Sebastian and Koschi, Markus and Althoff, Matthias},
  title     = {Worst-Case Analysis of the Time-To-React Using Reachable Sets},
  booktitle = {Proc. IEEE Intelligent Vehicles Symposium (IV)},
  address   = {Changshu, China},
  pages     = {1891--1897},
  year      = {2018},
  doi       = {10.1109/IVS.2018.8500709}
}

@inproceedings{sanchez2019parallel,
  author    = {S{\'a}nchez Morales, Eduardo and Membarth, Richard and Gaull, Andreas and Slusallek, Philipp and Dirndorfer, Tobias and Kammenhuber, Alexander and Lauer, Christian and Botsch, Michael},
  title     = {Parallel Multi-Hypothesis Algorithm for Criticality Estimation in Traffic and Collision Avoidance},
  booktitle = {Proc. IEEE Intelligent Vehicles Symposium (IV)},
  address   = {Paris, France},
  pages     = {2164--2171},
  year      = {2019},
  doi       = {10.1109/IVS.2019.8814015}
}

@article{xu2020integrated,
  author  = {Xu, Can and Zhao, Wanzhong and Wang, Chunyan},
  title   = {An Integrated Threat Assessment Algorithm for Decision-Making of Autonomous Driving Vehicles},
  journal = {IEEE Transactions on Intelligent Transportation Systems},
  volume  = {21},
  number  = {6},
  pages   = {2510--2521},
  year    = {2020},
  doi     = {10.1109/TITS.2019.2919865}
}

@misc{shalevshwartz2017formal,
  author       = {Shai Shalev{-}Shwartz and Shaked Shammah and Amnon Shashua},
  title        = {On a Formal Model of Safe and Scalable Self-driving Cars},
  journal      = {CoRR},
  volume       = {abs/1708.06374},
  year         = {2017},
  url          = {http://arxiv.org/abs/1708.06374},
  eprinttype   = {arXiv},
  eprint       = {1708.06374},
  bibsource    = {dblp computer science bibliography, https://dblp.org}
}

@article{kuhn1955hungarian,
  author  = {Kuhn, Harold W.},
  title   = {The {Hungarian} Method for the Assignment Problem},
  journal = {Naval Research Logistics Quarterly},
  volume  = {2},
  number  = {1-2},
  pages   = {83--97},
  year    = {1955},
  doi     = {10.1002/nav.3800020109}
}

@inproceedings{rusu20113d,
  author    = {Rusu, Radu Bogdan and Cousins, Steve},
  title     = {{3D} is here: {Point Cloud Library} ({PCL})},
  booktitle = {Proc. IEEE International Conference on Robotics and Automation (ICRA)},
  address   = {Shanghai, China},
  pages     = {1--4},
  year      = {2011},
  doi       = {10.1109/ICRA.2011.5980567}
}

@article{althoff2014online,
  author  = {Matthias Althoff and John M. Dolan},
  title   = {Online Verification of Automated Road Vehicles Using Reachability Analysis},
  journal = {IEEE Transactions on Robotics},
  year    = {2014},
  volume  = {30},
  number  = {4},
  pages   = {903--918},
  doi     = {10.1109/TRO.2014.2312453},
  url     = {https://ieeexplore.ieee.org/document/6784493}
}

@inproceedings{gruber2018anytime,
  author={Gruber, Felix and Althoff, Matthias},
  booktitle={2018 21st International Conference on Intelligent Transportation Systems (ITSC)}, 
  title={Anytime Safety Verification of Autonomous Vehicles}, 
  year={2018},
  volume={},
  number={},
  pages={1708-1714},
  doi={10.1109/ITSC.2018.8569950}}

@inproceedings{magdici2016failsafe,
  author={Magdici, Silvia and Althoff, Matthias},
  booktitle={2016 IEEE 19th International Conference on Intelligent Transportation Systems (ITSC)}, 
  title={Fail-safe motion planning of autonomous vehicles}, 
  year={2016},
  volume={},
  number={},
  pages={452-458},
  doi={10.1109/ITSC.2016.7795594}}

@inproceedings{schuermann2017drivability,
       booktitle = {ITSC 2017},
            year = {2017},
           title = {Ensuring Drivability of Planned Motions from Simple Models Using Formal Methods},
          author = {Sch{\"u}rmann, Bastian and He{\ss}, Daniel and Eilbrecht, Jan and Stursberg, Olaf and K{\"o}ster, Frank and Althoff, Matthias},
           month = {October},
             url = {https://elib.dlr.de/111912/}
}

@inproceedings{vomdorff2020failsafe,
  author={Dorff, Sebastian vom and Böddeker, Bert and Kneissl, Maximilian and Fränzle, Martin},
  booktitle={2020 Design, Automation \& Test in Europe Conference \& Exhibition (DATE)}, 
  title={A Fail-safe Architecture for Automated Driving}, 
  year={2020},
  volume={},
  number={},
  pages={828-833},
  doi={10.23919/DATE48585.2020.9116283}}

@article{hillenbrand2006multilevel,
  author  = {J{\"o}rg Hillenbrand and Andreas M. Spieker and Klaus Kroschel},
  title   = {A Multilevel Collision Mitigation Approach---Its Situation Assessment, Decision Making, and Performance Tradeoffs},
  journal = {IEEE Transactions on Intelligent Transportation Systems},
  year    = {2006},
  volume  = {7},
  number  = {4},
  pages   = {528--540},
  doi     = {10.1109/TITS.2006.883115},
  url     = {https://ieeexplore.ieee.org/document/4019437}
}

@article{eidehall2008statistical,
  author  = {Andreas Eidehall and Lars Petersson},
  title   = {Statistical Threat Assessment for General Road Scenes Using {Monte Carlo} Sampling},
  journal = {IEEE Transactions on Intelligent Transportation Systems},
  year    = {2008},
  volume  = {9},
  number  = {1},
  pages   = {137--147},
  doi     = {10.1109/TITS.2007.909241},
  url     = {https://ieeexplore.ieee.org/document/4414362}
}

@article{schreier2016integrated,
  author  = {Matthias Schreier and Volker Willert and J{\"u}rgen Adamy},
  title   = {An Integrated Approach to Maneuver-Based Trajectory Prediction and Criticality Assessment in Arbitrary Road Environments},
  journal = {IEEE Transactions on Intelligent Transportation Systems},
  year    = {2016},
  volume  = {17},
  number  = {10},
  pages   = {2751--2766},
  doi     = {10.1109/TITS.2016.2522507},
  url     = {https://ieeexplore.ieee.org/document/7412746}
}

@article{kim2018lane,
  author  = {Jaehwan Kim and Dongsuk Kum},
  title   = {Collision Risk Assessment Algorithm via Lane-Based Probabilistic Motion Prediction of Surrounding Vehicles},
  journal = {IEEE Transactions on Intelligent Transportation Systems},
  year    = {2018},
  volume  = {19},
  number  = {9},
  pages   = {2965--2976},
  doi     = {10.1109/TITS.2017.2768318},
  url     = {https://ieeexplore.ieee.org/document/8166818}
}

@article{Althoff2009Model-Based,
  title={Model-Based Probabilistic Collision Detection in Autonomous Driving},
  author={M. Althoff and O. Stursberg and M. Buss},
  journal={IEEE Transactions on Intelligent Transportation Systems},
  year={2009},
  volume={10},
  number={2},
  pages={299--310},
  doi={10.1109/TITS.2009.2018966}
}

@article{Lee2019Collision,
  title={Collision Avoidance/Mitigation System: Motion Planning of Autonomous Vehicle via Predictive Occupancy Map},
  author={Kibeom Lee and Dongsuk Kum},
  journal={IEEE Access},
  year={2019},
  volume={7},
  pages={52846--52857},
  doi={10.1109/ACCESS.2019.2912067}
}

@inproceedings{Shen2020Collision-Free,
  title={Collision-Free Path Planning for Automated Vehicles Risk Assessment via Predictive Occupancy Map},
  author={Dan Shen and Yaobin Chen and Lingxi Li and Stanley Chien},
  booktitle={2020 IEEE Intelligent Vehicles Symposium (IV)},
  year={2020},
  pages={985--991},
  doi={10.1109/IV47402.2020.9304720}
}

@article{Lee2019Study,
  title={Study on the Improvement of a Collision Avoidance System for Curves},
  author={JungEun Lee and GyoungEun Kim and ByeongWoo Kim},
  journal={Applied Sciences},
  year={2019},
  volume={9},
  number={24},
  pages={5380},
  doi={10.3390/app9245380}
}

@article{Han2025Adaptive,
  title={Adaptive ROI for Collision Warning Mitigation Based on Road Geometry and Kinematics},
  author={Seungho Han and Minseong Choi and Byeonggwan Jang and Keun Ha Choi and Kyung-Soo Kim},
  journal={IEEE Sensors Journal},
  year={2025},
  volume={25},
  number={3},
  pages={5499--5509},
  doi={10.1109/JSEN.2024.3512947}
}

@article{Gao2022Deformable,
  title={Deformable Radar Polygon: A Lightweight and Predictable Occupancy Representation for Short-Range Collision Avoidance},
  author={Xiangyu Gao and Sihao Ding and Harshavardhan Reddy Dasari},
  journal={IEEE Sensors Journal},
  year={2024},
  volume={24},
  pages={26104--26114},
  doi={10.1109/JSEN.2024.3422148}
}

@article{Shivanna2024Object,
  title={Object Detection, Recognition, and Tracking Algorithms for ADASs---A Study on Recent Trends},
  author={Vinay Malligere Shivanna and Jiun-In Guo},
  journal={Sensors},
  year={2023},
  volume={24},
  number={1},
  pages={249},
  doi={10.3390/s24010249}
}

@article{Abrecht2024Deep,
  title={Deep Learning Safety Concerns in Automated Driving Perception},
  author={Stephanie Abrecht and Alexander Hirsch and Shervin Raafatnia and Matthias Woehrle},
  journal={IEEE Transactions on Intelligent Vehicles},
  year={2025},
  volume={10},
  number={2},
  pages={1455--1466},
  doi={10.1109/TIV.2024.3428415}
}

@article{zablocki2022explainability,
  title={Explainability of Deep Vision-Based Autonomous Driving Systems: Review and Challenges},
  author={Zablocki, {\'E}loi and Ben-Younes, H{\'e}di and P{\'e}rez, Patrick and Cord, Matthieu},
  journal={International Journal of Computer Vision},
  year={2022},
  volume={130},
  number={10},
  pages={2425--2452},
  doi={10.1007/s11263-022-01657-x}
}

@article{Peng2024Curse,
  title={Curse of Rarity for Autonomous Vehicles},
  author={Liu, Henry X. and Feng, Shuo},
  journal={Nature Communications},
  year={2024},
  volume={15},
  number={1},
  doi={10.1038/s41467-024-49194-0}
}

\end{document}